\newif\ifCameraReady
\CameraReadytrue 

\documentclass[letterpaper, 10 pt, conference]{ieeeconf}
\ifCameraReady
    \IEEEoverridecommandlockouts
\fi
\usepackage{graphicx} 
\usepackage{ragged2e} 
\usepackage{array}
\usepackage{multirow}
\usepackage{booktabs} 
\usepackage{url}
\usepackage{xcolor}
\usepackage{pifont}
\usepackage{dirtree}

\usepackage{cite}
\usepackage[hidelinks]{hyperref}
\hypersetup{
    colorlinks=true,
    linkcolor=blue,
    citecolor=blue,
    urlcolor=blue,
    pdfstartview=FitH,
    pdfauthor={Your Name},
}
\usepackage{colortbl} 
\usepackage{diagbox}

\usepackage{pgfplots}
\usepgfplotslibrary{groupplots}
\pgfplotsset{compat=1.18}
\newcommand{\cmark}{\textcolor[rgb]{0,0.5,0}{\ding{51}}} 
\newcommand{\xmark}{\textcolor[rgb]{0.6,0,0}{\ding{55}}} 

\title{\LARGE \bf
    CU-Multi: A Dataset for Multi-Robot Collaborative Perception
}

\ifCameraReady
    \author{Doncey Albin$^{1}$, Daniel McGann$^{2}$, Miles Mena$^{1}$, Annika Thomas$^{3}$, Harel Biggie$^{4}$, Xuefei Sun$^{1}$\\ Steve McGuire$^{5}$, Jonathan P. How$^{3}$, and Christoffer Heckman$^{1}$%
    \thanks{$^{1}$Authors are with the Autonomous Robotics and Perception Group at the University of Colorado Boulder, Boulder, CO 80309, USA}%
    \thanks{$^{2}$Daniel McGann with the Robot Perception Lab at Carnegie Mellon University, Pittsburgh, PA 15213, USA}%
    \thanks{$^{3}$Authors are with the Aerospace Controls Laboratory at Massachusetts Institute of Technology, Cambridge, MA 02139, USA}%
    \thanks{$^{4}$Harel Biggie with the Computer Science and Artificial Intelligence Laboratory at Massachusetts Institute of Technology, Cambridge, MA 02139, USA}%
    \thanks{$^{5}$Steve McGuire with the Human-Aware Robotic Exploration Lab at University of California Santa Cruz, Santa Cruz, CA 95064, USA}%
    \thanks{This work was supported in part by USDA-NIFA award 2021-67021-33450 and by the National Science Foundation (NSF) award \#2339328.}}
\fi

\ifCameraReady
  \newcommand{\dataseturl}{\url{https://github.com/arpg/CU-Multi}}
\else
  \newcommand{\dataseturl}{https://github.com/cumultidataset/CU-Multi}
\fi

\begin{document}

\maketitle
\thispagestyle{empty}
\pagestyle{empty}

\begin{abstract}
A central challenge for multi-robot systems is fusing independently gathered perception data into a unified representation. Despite progress in Collaborative SLAM (C-SLAM), benchmarking remains hindered by the scarcity of dedicated multi-robot datasets. Many evaluations instead partition single-robot trajectories, a practice that may only partially reflect true multi-robot operations and, more critically, lacks standardization, leading to results that are difficult to interpret or compare across studies. While several multi-robot datasets have recently been introduced, they mostly contain short trajectories with limited inter-robot overlap and sparse intra-robot loop closures. To overcome these limitations, we introduce CU-Multi, a dataset collected over multiple days at two large outdoor sites on the University of Colorado Boulder campus. CU-Multi comprises four synchronized runs with aligned start times and controlled trajectory overlap, replicating the distinct perspectives of a robot team. It includes RGB-D sensing, RTK GPS, semantic LiDAR, and refined ground-truth odometry. By combining overlap variation with dense semantic annotations, CU-Multi provides a strong foundation for reproducible evaluation in multi-robot collaborative perception tasks. The dataset, support code, and updates are available at \dataseturl.
\end{abstract}

\begin{figure}[thpb]
  \centering
  \includegraphics[width=0.95\linewidth]{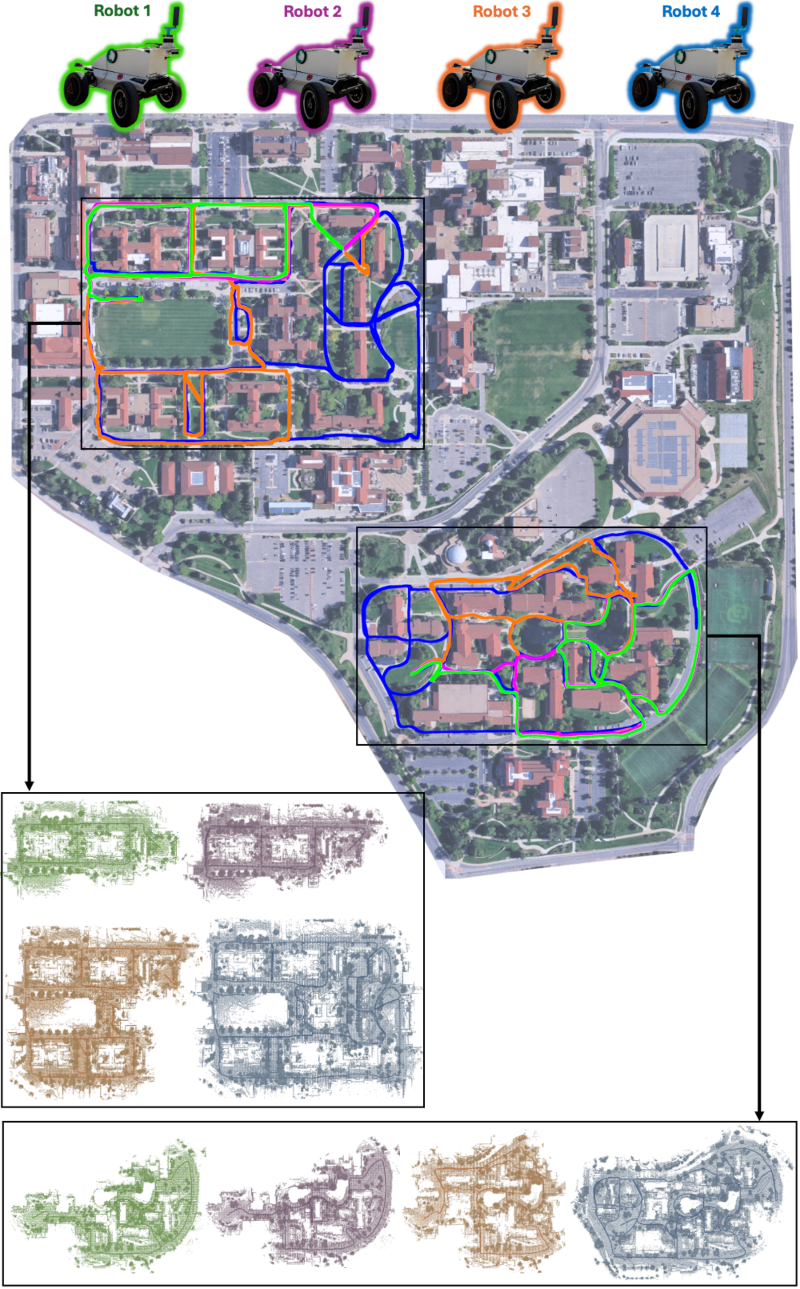}
  \caption{Overhead view of all paths overlaid on map from the Main Campus (top) and Kittredge Loop (bottom) environments in the CU-Multi Dataset.}
  \label{fig:main}
\end{figure}
\begin{table*}
    \caption{Comparison of field-collected multi-robot datasets. Note: only datasets with LiDAR have been listed. }
    \label{tab:multi_robot_datasets}
    \centering
    \begin{tabular}{
    >{\RaggedRight}m{2.2cm} 
    >{\RaggedRight}m{1.5cm} 
    >{\RaggedRight}m{2.8cm} 
    >{\centering\arraybackslash}m{1.7cm} 
    >{\centering\arraybackslash}m{1.0cm} 
    >{\centering\arraybackslash}m{0.6cm} 
    >{\centering\arraybackslash}m{0.6cm} 
    >{\centering\arraybackslash}m{1.0cm} 
    >{\centering\arraybackslash}m{1.5cm} 
    } 
        \toprule
        \textbf{Dataset} & \textbf{\# Robots} & \textbf{Environments} & \textbf{Multi-Session} & \textbf{RGB-D} & \textbf{GPS} & \textbf{IMU} & \textbf{Annotated LiDAR} & \textbf{Longest Trajectory (km)} \\ [0.5ex]
        
        \midrule 
        
        GRACO \cite{zhu2023graco} & 1 GW, 2 A & 8 Out & \xmark & \cmark & \cmark & \cmark & \xmark & 0.82 \\
        
        Kimera-Multi \cite{Tian23iros_KimeraMultiExperiments} & 8 GW & 1 Out, 1 In, 1 Hybrid & \xmark & \cmark & \xmark & \cmark & \xmark & 1.40 \\
        
        S3E  \cite{feng2024s3e} & 3 GW & 13 Out, 5 In & \xmark & \cmark & \cmark & \cmark & \xmark & 1.95 \\
        
        DiTer++ \cite{kim2024diter++} & 2 GL & 3 Out & \cmark & \cmark & \xmark & \cmark & \xmark & N/A \\
        
        Lamp 2.0 \cite{chang2022lamp} & 4 GW, 3 GL & 1 Out, 3 Sub & \xmark & \cmark & \xmark & \cmark & \xmark & 2.20 \\
        
        CoPeD \cite{zhou2024coped} & 3 GW, 2 A & 3 In, 3 Out & \xmark & \cmark & \cmark & \cmark & \xmark & 0.36 \\
        
        \textbf{CU-Multi (ours)} & \textbf{4 GW} & \textbf{2 Out} & \cmark & \cmark & \cmark & \cmark & \cmark & \textbf{4.01} \\[0.5ex]
        \bottomrule
    \end{tabular}
    \\[0.5ex]
    \footnotesize{\textbf{GW}, \textbf{GL}, and \textbf{A} denote ground-wheeled, ground-legged, and aerial  robots, respectively. \textbf{In}, \textbf{Out}, and \textbf{Hybrid} represent indoor, outdoor, and a mix of both, respectively.}
\end{table*}
\section{Introduction}
Multi-robot systems significantly enhance capabilities across diverse domains, particularly in large-scale environments, by accelerating exploration through distributed sensing and collaborative decision-making \cite{kim2023multi}. A central challenge in realizing these advantages lies in fusing perception data collected independently by multiple robots into a unified global representation, complicated by spatial and temporal misalignment. This challenge is further compounded in many practical scenarios where external positioning methods (e.g. GPS, motion capture) are impractical, unreliable, or hazardous. Multi-robot Collaborative SLAM (C-SLAM) algorithms are typically first developed and validated offline using datasets before doing field tests in real-world conditions, where system-level issues often arise. Consequently, the availability of realistic, well-structured multi-robot datasets is essential for supporting reproducible research and bridging the gap between offline development and real-world deployment.

Several multi-robot datasets have recently been introduced for C-SLAM verification \cite{feng2024s3e, zhou2024coped, zhu2023graco}. These datasets include synchronized multi-agent sensor data collected across both indoor and/or outdoor environments. While these datasets represent strong starting points, there remains a need for longer-trajectory multi-robot datasets with explicitly varied overlap (see Table~\ref{tab:multi_robot_datasets}). Despite recent advancements in multi-robot datasets, there remains a prevalent practice of artificially segmenting a single trajectory into multiple parts to simulate a multi-robot scenario for verification \cite{zhong2023dcl}. Single-robot SLAM datasets still offer a few benefits over many of the existing multi-robot datasets, such as large trajectories with multiple loop closures, as well as the availability of both camera and LiDAR semantics \cite{zhong2023dcl, lajoie2023swarm}. However, without careful consideration, arbitrary segmentation may not accurately represent realistic observational overlap typically encountered in multi-robot operations \cite{lajoie2022towards}. 

In this paper, we introduce CU-Multi, a dataset designed to support the evaluation of methods relevant to multi-robot perception. These methods include C-SLAM, LiDAR and visual data association, and multi-session LiDAR-based place recognition. CU-Multi offers the following key features:
\begin{enumerate}
    \item \textit{A multi-robot dataset} consisting of two large-scale environments, each with four robots. Our dataset contains a total of eight diverse trajectories, spanning a combined length of 16.7 km across the CU Boulder campus (Figure~\ref{fig:main}).
    \item \textit{Systematically varied trajectory overlaps and a rendezvous-based trajectory design}, enabling evaluation of multi-robot perception tasks under different levels of observational redundancy and variation.
    \item \textit{Refined ground truth poses} using RTK GPS, a digital elevation model, LiDAR-inertial SLAM, and a highly accurate scan-matching algorithm, calculated at each LiDAR timestamp.
    \item \textit{Semantic labels} for all LiDAR scans, dataset tools for interacting with the data, as well as initial benchmarks on a well-known C-SLAM and LiDAR place recognition method to demonstrate dataset usability and utility of the provided tools.
\end{enumerate}

\section{Related Work}
Although single-robot SLAM has long benefited from curated datasets, extending SLAM to multi-robot systems introduces challenges that demand more representative data of a true multi-robot system. A common practice is to simulate multi-robot scenarios by splitting single-robot trajectories, but this raises concerns about realism and consistency. This section surveys single-robot datasets, the trajectory-splitting strategies applied to them, and existing multi-robot datasets, motivating the need for our proposed contribution.

\subsection{Single-Robot SLAM Datasets}
One of the most widely used datasets for evaluating single-robot SLAM and global localization pipelines is the KITTI Odometry Benchmark \cite{geiger2013vision}, a large-scale computer vision dataset collected in Karlsruhe, Germany using a sensor-equipped automobile. It features stereo imagery, LiDAR, and GPS/IMU data across 22 driving sequences in urban and rural environments. In 2019, SemanticKITTI \cite{behley2019semantickitti} expanded the original KITTI dataset to include dense point-wise annotations for all LiDAR scans, enabling tasks such as semantic segmentation and scene completion. KITTI-360 \cite{liao2022kitti}, released in 2022, was also collected in Karlsruhe and contains broader city coverage, improved GPS accuracy, and consistent 2D/3D semantic instance annotations across panoramic imagery and LiDAR scans. Although originally intended for single-robot perception tasks, these datasets are commonly repurposed for verifying and benchmarking multi-robot C-SLAM pipelines through segmenting a single trajectory into multiple parts to simulate independent robots.

\subsection{Trajectory Splitting on Single-Robot Datasets}
 Early work by Cieslewski and Scaramuzza demonstrated the approach of using a single-robot dataset for C-SLAM \cite{cieslewski2017efficient, cieslewski2018data}, and subsequent methods including DOOR-SLAM \cite{lajoie2020door}, DiSCO-SLAM \cite{huang2021disco}, and RDC-SLAM \cite{xie2021rdc} adopted similar practices. More recent systems, such as DCL-SLAM \cite{zhong2023dcl}, Wi-Closure \cite{wang2023wi}, and SWARM-SLAM \cite{lajoie2023swarm}, continue to rely on partitioning KITTI or KITTI-360 sequences, often supplemented with small multi-robot datasets like GRACO \cite{zhu2023graco} or S3E \cite{feng2024s3e}.

While splitting provides convenient large-scale data, its limitations are well documented.  Lajoie et al.\  cautioned that partitioning trajectories at overlapping regions creates unrealistic assumptions, as identical viewpoints and lighting are unlikely to occur simultaneously in true multi-robot operations \cite{lajoie2022towards}. Despite this, most works do not specify their partitioning strategies, and no standardized protocol exists. As seen in methods such as FRAME \cite{stathoulopoulos2024frame}, SideSLAM \cite{liu2024slideslam}, and Cao et al.\ \cite{cao2024multi}, overlapping splits can even reuse the same observations, undermining realism. This persistent reliance highlights a key gap: the lack of standardized datasets that capture the diversity and independence of multi-robot observations.

\subsection{Multi-Robot Datasets}
One of the earliest multi-robot datasets is the 2011 UTIAS dataset \cite{leung2011utias}, which includes nine indoor environments, each featuring five wheeled robots equipped with monocular cameras. In 2020, two additional datasets were introduced: FordAV \cite{agarwal2020ford} and AirMuseum \cite{dubois2020airmuseum}. FordAV provides multi-seasonal data collected by a fleet of Ford Fusion vehicles outfitted with four LiDARs, GPS, and six cameras. However, the vehicle trajectories exhibit near-complete overlap across all sequences, requiring users to manually segment them to simulate limited overlap scenarios. AirMuseum captures warehouse-scale data using a heterogeneous team of two ground robots and one aerial platform, each equipped with a stereo camera rig. While both AirMuseum and UTIAS offer multi-robot observations, their indoor collection settings and lack of LiDAR data limit their scalability and applicability to modern perception pipelines.

More recent datasets have broadened the scope of multi-robot data collection. GRACO \cite{zhu2023graco} introduced a heterogeneous two-robot team consisting of a ground and aerial robot; however, the dataset is constrained by a low-resolution 16-beam LiDAR and short trajectories. In follow-up work to Kimera-Multi \cite{tian2022kimera}, Kim et al.\ released a dataset comprising three sequences of eight ground robots operating in indoor, outdoor, and hybrid environments \cite{Tian23iros_KimeraMultiExperiments}. The CoPeD dataset goes further by addressing specific limitations of prior datasets in terms of sensor heterogeneity and environmental diversity \cite{zhou2024coped}. Kimera-Multi and CoPeD address valuable aspects needed for a multi-robot dataset, but are still limited to small trajectories and minimal observational overlap.

The need for datasets that explicitly explore varying degrees of shared observations has also gained attention. S3E presents four distinct trajectory paradigms to study different levels of inter-robot trajectory overlap using a team of three robots \cite{feng2024s3e}. In their work, they intended to reduce the amount of inter-robot trajectory overlap to simulate scenarios that are not favorable. While this approach is a valuable step toward evaluating C-SLAM methods under varied trajectory overlap conditions, the trajectories remain limited in scale and contain relatively few intra-robot loop closures.

Despite recent progress in multi-robot datasets, many works have yet to adopt them for evaluation. Relying on idealized, single-trajectory conditions can obscure failure modes that emerge when robots operate independently, with varied trajectory overlap, differing viewpoints, and asynchronous data capture. As research in multi-robot perception continues to mature, there is a growing need for datasets that provide precise ground truth across multiple independently moving platforms to rigorously evaluate C-SLAM methods and inter-robot loop closure detection. Our dataset addresses this gap by offering multi-robot trajectories with explicitly controlled overlap levels and annotated LiDAR scans, enabling more realistic and robust evaluation of multi-robot perception pipelines.

\begin{figure}[ht]
  \centering
  \includegraphics[width=0.98\linewidth]{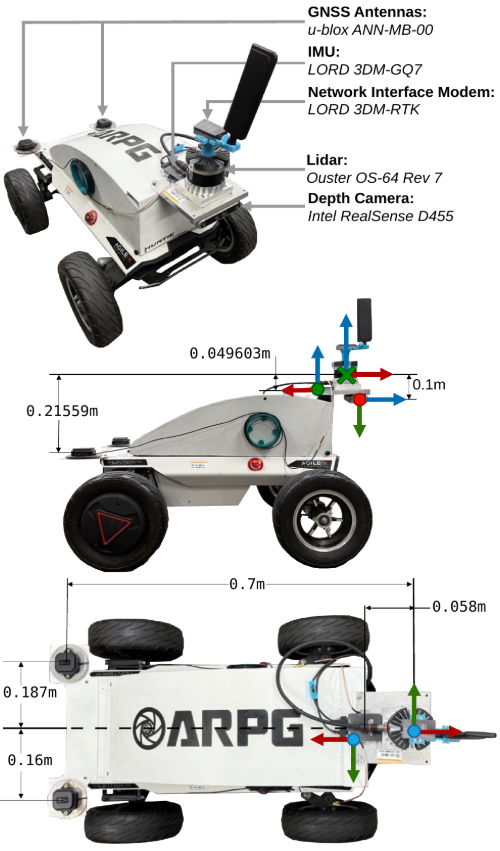}
  \caption{Hardware and sensor specifications used on platform.}
  \label{fig:sensors}
\end{figure}
\begin{figure*}[t]
  \centering
  \includegraphics[width=0.98\linewidth]{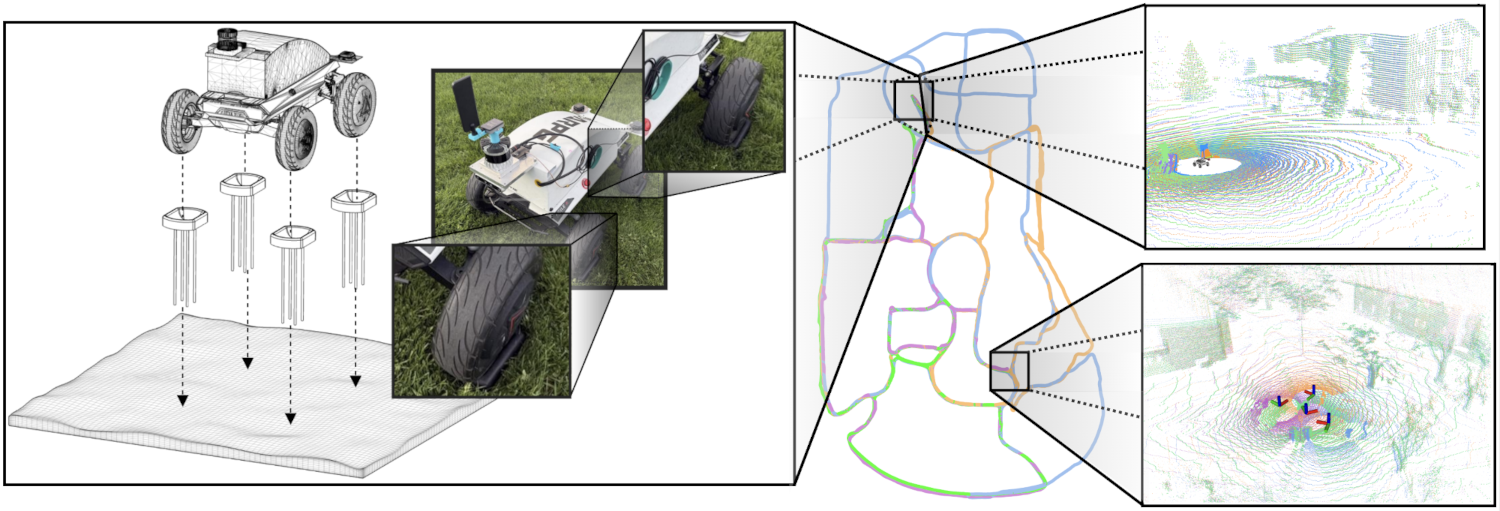}
  \caption{During data collection, we ensured each session started from the same initial starting point for each environment through securing our platform onto four wheel chocks that were staked into grass (left). During data playback, we can see the results of our initial alignment through tightly overlapping lidar scans (top right). Lastly, we ended each session at a common location, each run ending within 5m of one another (bottom right).}
  \label{fig:data_collection}
\end{figure*}
\section{The  CU-Multi Dataset}
The CU-Multi dataset was collected using a single robotic platform over multiple sessions. This section describes the platform and sensors used for data collection, our novel approach to ensuring accurate ground-truth localization across multiple runs within each environment, our two-stage method for LiDAR annotation, the large-scale outdoor environments where data was gathered, and how the dataset is structured.
    
\subsection{Platform and Sensors}
The CU-Multi dataset was collected using the AgileX Hunter SE platform modified with a custom-designed electronics housing (see Figure~\ref{fig:sensors}). Inside the housing, the system includes an Intel NUC i7 with 16 GB of RAM, a power distribution board, and a 217 Wh battery. Externally, a mounting plate on the rear of the Hunter SE accommodates two u-blox ANN-MB-00 GNSS antennas, while a front-mounted plate holds a 64-beam Ouster LiDAR sensor, a RealSense D455 RGB-D camera, a Lord MicroStrain G7 IMU, and a Lord RTK cellular modem with an external antenna. To ensure precise odometry estimation, LiDAR timestamps are hardware synchronized with the onboard computer's timeclock using Precision Time Protocol (PTP).

\subsection{Dataset Collection and Environments}
The CU-Multi dataset was collected across two large outdoor environments on the University of Colorado Boulder campus. The first environment, \textit{main\_campus}, covers the central academic area, with approximately 7.4 km of traversed path. The second environment, \textit{kittredge\_loop}, encompasses a large region south of the main campus, offering more open space and varied terrain for multi-robot exploration.

Each of the two environments consists of four trajectories with varying levels of overlap. In our dataset, \emph{robot1} and \emph{robot2} share significant path overlap but capture the scene from different viewpoints, enabling users to test algorithms for robust multi-session data association. Meanwhile, \emph{robot3} and \emph{robot4} follow paths with less overlap, providing opportunities to evaluate performance under sparse or partially overlapping observations. The trajectory of \emph{robot3} extends the trajectory of the first two robots, introducing additional observations. Finally, \emph{robot4} encompasses all previous trajectories, covering the most extensive area. Formally, the relationship between these trajectories can be expressed as:
\begin{equation}
    T_1 \approx T_2, \quad (T_1 \cup T_2) \subseteq T_3, \quad (T_1 \cup T_2 \cup T_3) \subseteq T_4
 \end{equation}
where \( T_i \) represents the trajectory of $\emph{robot}_i$.

To maximize viewpoint diversity, we introduced variations in observational perspectives while traversing overlapping regions. The structured overlap across trajectories enables users to select data subsets based on their desired level of spatial redundancy and cross-view consistency. All trajectories end within a distance of 5 meters of each other at a common location in each environment, representing a multi-robot rendezvous scenario (Figure~\ref{fig:data_collection}).

\begin{figure}[th]
  \centering
  \includegraphics[width=\linewidth]{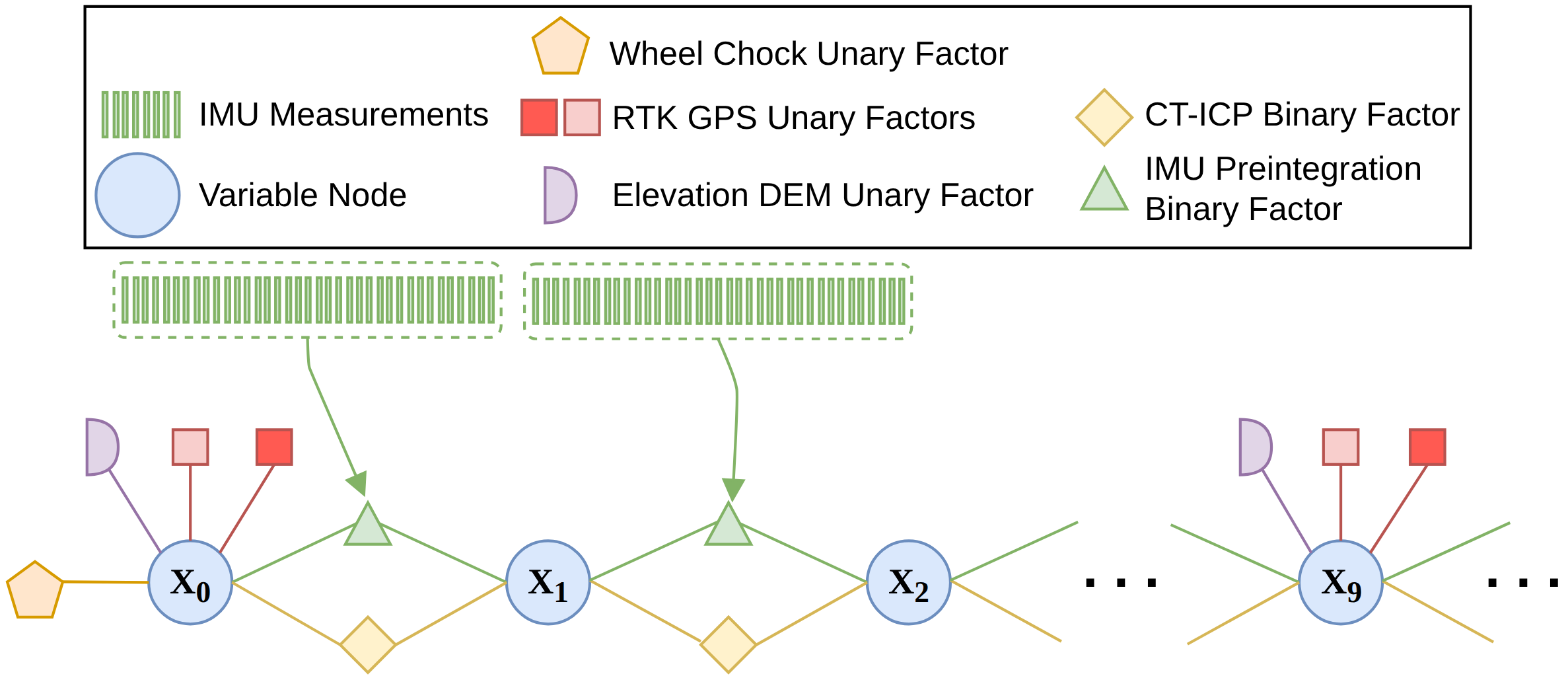}
  \caption{The factor-graph used to generate ground truth for all trajectories. The LiDAR and IMU factors interpolate between the poses that are well constrained by external reference data (i.e. GPS, Elevation, Chock Prior).}
  \label{fig:groundtruth_factorgraph}
\end{figure}
\begin{figure}[t]
  \centering
  \includegraphics[height=1.2   in]{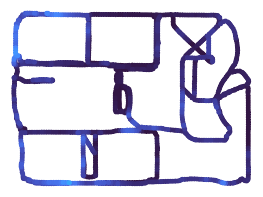}
  \includegraphics[trim={0 0.3cm 0 0.4cm},clip,height=1in]{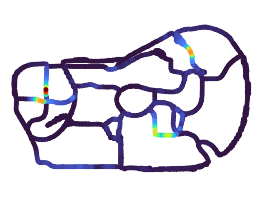}
  \caption{The GPS uncertainty in both data collection environments reported as $||\Sigma||_F$. Covariances ($\Sigma$) are reported by the GNSS and are spatially averaged across all trajectories using a 4m resolution voxel-grid.}
  \label{fig:gps_uncertainty}
\end{figure}

\subsection{Ground Truth}
\label{sec:gto_geospatial_alignment}
To support quantitative evaluation, we provide globally aligned ground truth for all trajectories. We initialized each trajectory from a fixed point by securing the Hunter SE platform onto wheel chocks staked into the grass (Figure~\ref{fig:data_collection}), after which the robot remained stationary for $\sim 60$s to acquire an accurate RTK GPS fix. However, due to the density of buildings in traversed sections of the environments, the RTK GPS does lose some accuracy, particularly in elevation (see Figure~\ref{fig:gps_uncertainty}), and at 2Hz the GPS data alone is too sparse to provide useful ground truth.

\begin{figure}[bh]
  \centering
  \includegraphics[width=\linewidth]{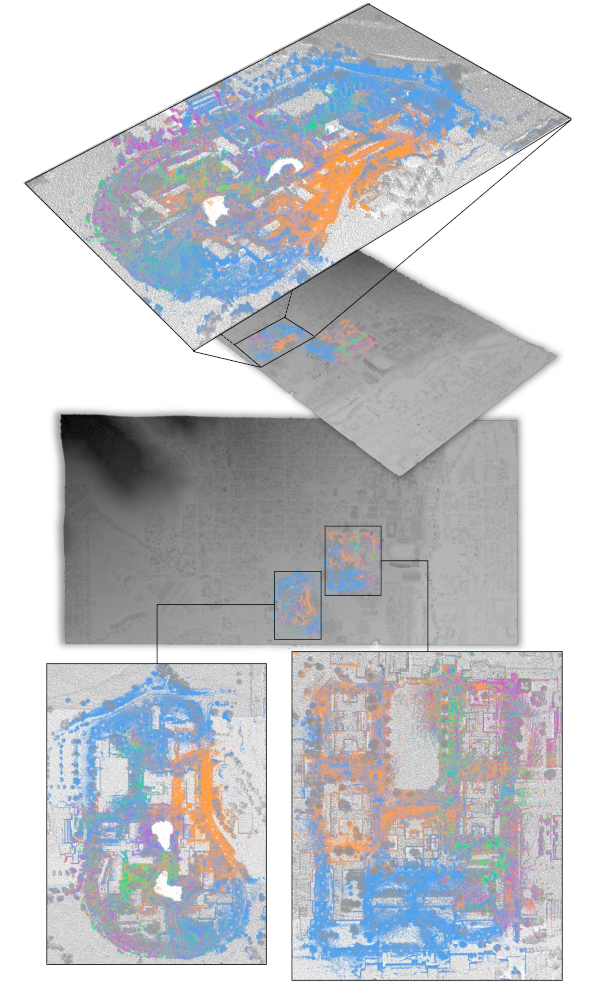}
  \caption{Maps constructed from the estimated ground truth poses overlaid on an external DSM for qualitative validation. Maps are colored per robot. Top: Kittredge Loop (perspective). Bottom: Main Campus (left) and Kittredge Loop (right) in overhead view.}
  \label{fig:groundtruth_maps}
\end{figure}

To address these limitations, we formulate a factor-graph incorporating GPS measurements from both GNSS receivers, 20Hz LiDAR odometry factors from LIO-SAM~\cite{liosam2020shan} and CT-ICP~\cite{dellenbach2022cticp}, and IMU preintegration factors to interpolate between GPS fixes. Elevation priors from a digital elevation model (DEM)\footnote{We use the 1m-resolution DEM of Boulder, CO provided by the USGS~\cite{usgs2020codem}.} mitigate large elevation ambiguity in dense areas, and a unary prior at the wheel chock location constrains each trajectory's initial pose. The resulting graph is optimized using \texttt{gtsam}~\cite{dellaert2012gtsam}, yielding a high-quality pose estimate at every LiDAR scan timestamp. We qualitatively validate these estimates by overlaying the reconstructed maps onto an independently acquired digital surface model (DSM)\footnote{Derived from the DRCOG 2020 QL1 LiDAR collection~\cite{drcog2020lidar}.} collected over the same area, observing strong spatial agreement across both environments (Fig.~\ref{fig:groundtruth_maps}).

\begin{figure}[t]
  \centering
  \includegraphics[width=\linewidth]{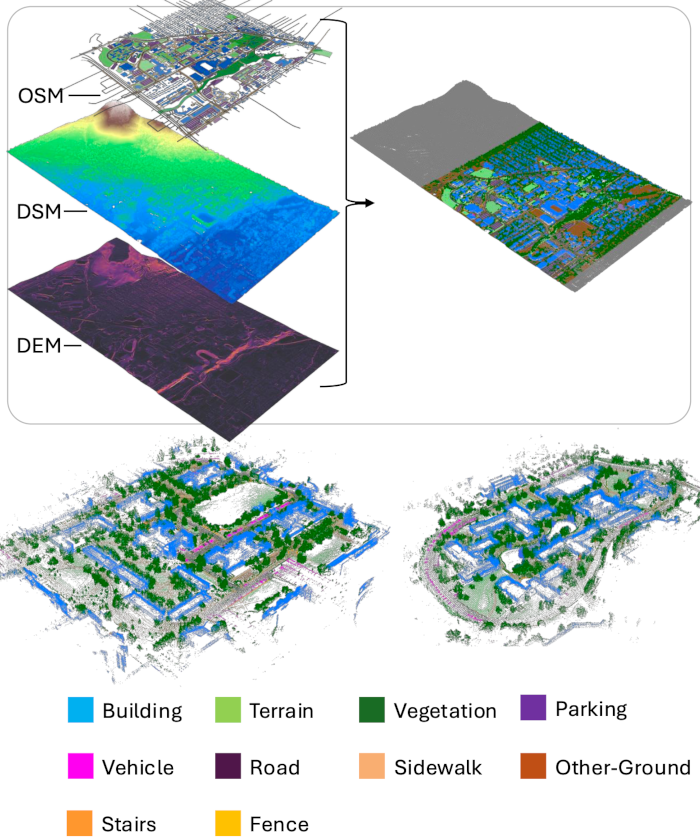}
      \caption{Illustration of the automated semantic labeling pipeline. External geospatial data sources, including OpenStreetMap (OSM) features, a digital surface model (DSM), and a digital elevation model (DEM), are fused to construct a global semantic map (top). This semantic prior is then used to generate semantically annotated LiDAR scans, which are aggregated into semantic environment maps for the Main Campus (bottom left) and Kittredge Loop (bottom right) sequences.}
  \label{fig:3d_semantics_aligned}
\end{figure}


\subsection{LiDAR Annotation}
\label{sec:lidar_annotations}
None of the multi-robot datasets surveyed in this work provide ground-truth semantic annotations for LiDAR data, although CoPeD~\cite{zhou2024coped} does include semantic labels for images. Generating LiDAR semantic annotations at scale is expensive and remains a major obstacle to building such datasets. To address this gap, we develop an automated labeling pipeline that combines accurate ground-truth odometry with open geospatial data, including the DEM and DSM described in Section~\ref{sec:gto_geospatial_alignment}, as well as OpenStreetMap (OSM) features (Fig.~\ref{fig:3d_semantics_aligned}).

First, we use the DEM to identify which DSM points belong to the ground surface, complementing the limited class information available in the DSM itself. We then use semantic OSM features to label the DSM, producing a semantic DSM whose annotations are constrained by whether points correspond to ground-level regions or above-ground structures. Finally, we transfer these labels to each LiDAR scan by assigning to every LiDAR point the label of its nearest neighbor in the semantic DSM. This procedure is enabled by the globally aligned, accurate ground-truth odometry, which allows reliable registration between the sensor data and the geospatial priors.

\begin{figure*}[t]
  \centering
  \includegraphics[width=0.95\linewidth]{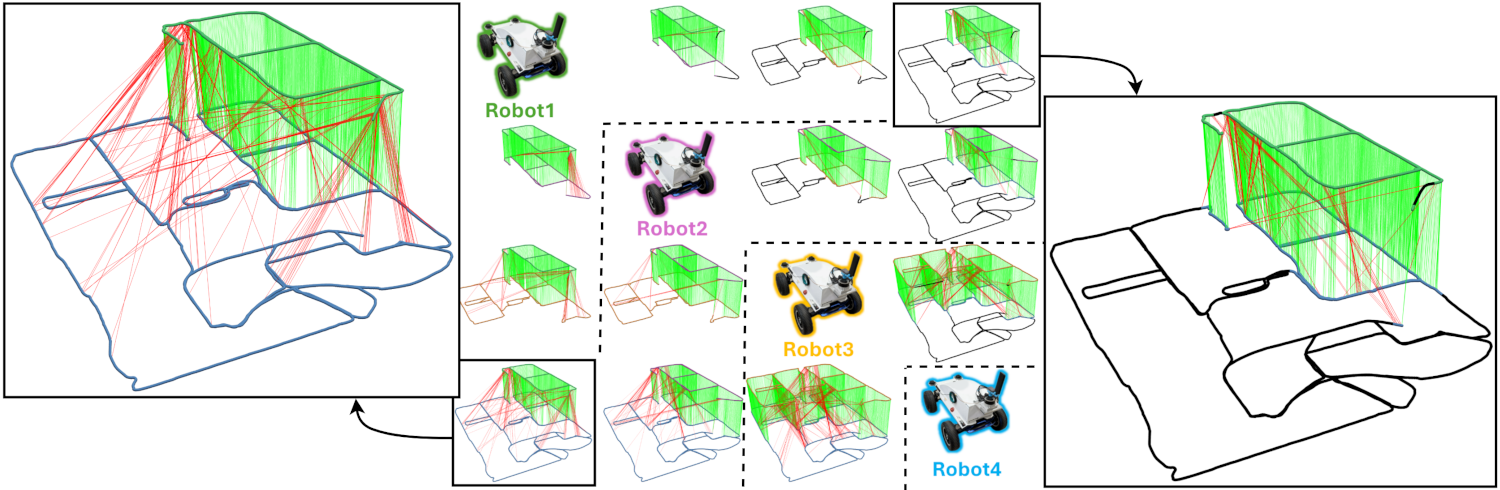}
  \caption{Example demonstrating utilization of LiDAR place recognition using ScanContext on data collected in Main Campus environment. All examples above the main diagonal compare multi-session place recognition and all examples below are multi-robot comparisons. Red lines show false matching loops and green lines show correct matching loops. We determine a loop-closure is correct if the positions of the scans are within 10m.} 
  \label{fig:LPR_experiment}
\end{figure*}

\subsection{Dataset Structure and Format}
The dataset is structured hierarchically, with environments and a calibration directory at the top level. The calibration directory (\textit{calib}) contains the relative sensor positions on the platform, camera intrinsics, a mesh and URDF of the system. Within each environment directory, there are four subdirectories corresponding to individual robots, labeled \textit{robot1} through \textit{robot4}.

Each robot-specific directory follows a consistent structure with subdirectories containing ROS2 \texttt{.db3} bags for the onboard RGB-D camera, LiDAR, IMU, GPS, and ground truth poses. The ground truth poses in the ROS2 \texttt{.db3} bags are centered about the starting position, therefore we also include a \texttt{.csv} in each robot subdirectory with the ground truth poses in UTM coordinates with each line formatted as \texttt{[timestamp, x, y, z, qx, qy, qz, qw]}. Lastly, on our project website, we illustrate the topics available to users in each of the ROS2 bags, which sensor (if applicable) collected the topic data, the rate of that topic, and the ROS2 message type in which the corresponding data are stored.


\subsection{Dataset Tools}
To facilitate adoption and reproducible experimentation, CU-Multi is released with a suite of tools. These include Python scripts for converting the raw data into common formats such as ROS1 bag files and ROS2 \texttt{.mcap} files. We also provide an interactive trajectory clipper tool that allows users to select, crop, and replay robot trajectories. This enables controlled-overlap scenarios beyond those observed during data collection, while still using runs that were independently captured with varied viewpoints. Such functionality is particularly useful for evaluating C-SLAM methods that employ distance-based communication constraints, where playback overlap can ensure that robot observations occur within a desired proximity. Together, these tools streamline integration with existing multi-robot perception pipelines, enabling rapid and flexible evaluation. Documentation and usage examples are provided on the project website.

\section{Use Case Demonstrations}
To demonstrate CU-Multi’s suitability for multi-robot C-SLAM and data association, we present two illustrative evaluations. For this purpose, we select two lightweight baselines: the LiDAR global descriptor ScanContext~\cite{kim2018scan} and the distributed C-SLAM system DiSCo-SLAM. Since DiSCo-SLAM employs ScanContext for inter-robot loop closure detection, we apply identical ScanContext parameters in both evaluations to ensure consistency. These parameters, along with step-by-step instructions, are released with our code to support reproducibility and facilitate drop-in integration of future methods.

\subsection{LiDAR Place Recognition}
ScanContext constructs a rotation-tolerant global descriptor for each LiDAR scan to enable fast retrieval. We evaluate precision-recall in both multi-session and multi-robot settings for each CU-Multi environment. In the multi-session setting, we compare descriptors only between pose pairs whose ground-truth separation is at most 10m across sessions. In the multi-robot setting, we perform all-pairs retrieval: all scans from one robot are compared against all scans from another robot (Fig.~\ref{fig:LPR_experiment}). We provide PR curves for both the multi-robot and multi-session place recognition experiments (Fig.~\ref{fig:LPR_precision_recall}).

\begin{figure*}[ht]
  \centering
  \includegraphics[width=0.95\linewidth]{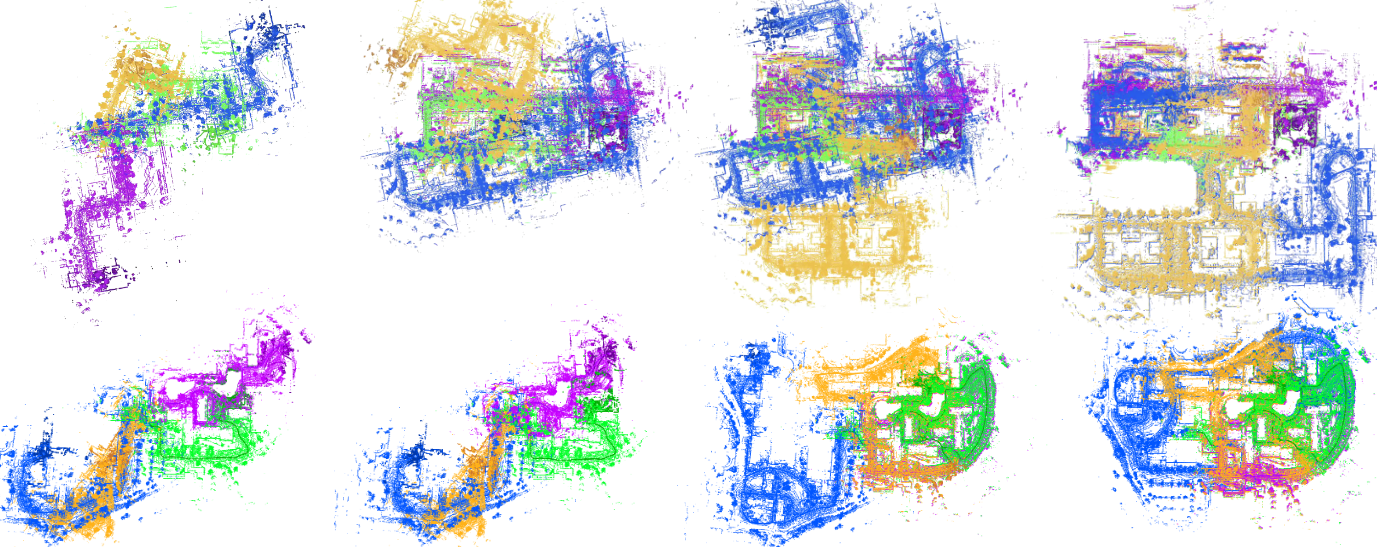}
  \caption{\textit{DiSCo-SLAM on CU-Multi.} \textbf{Top row}: Main Campus environment. \textbf{Bottom row}: Kittredge Loop environment. Each row illustrates four stages of alignment from left to right: (1) initial unaligned robot trajectories, (2) \textit{robot2} aligned to \textit{robot1}, (3) \textit{robot3} subsequently aligned with \textit{robot1} and \textit{robot2}, and (4) \textit{robot4} aligned with the remaining robots, resulting in all trajectories progressively unified into a common frame.}
  \label{fig:discoslam_full}
\end{figure*}
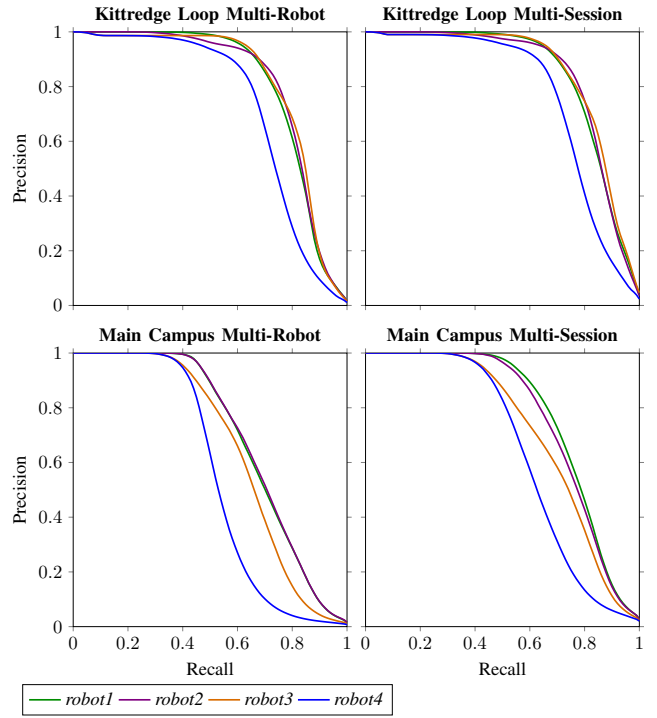
\begin{figure}[t]
  \centering
  \resizebox{\linewidth}{!}{%
\begin{tikzpicture}[outer sep=0pt]
  \pgfplotsset{
    baseaxis/.style={
      xmin=0, xmax=1, ymin=0, ymax=1,
      xtick={0,0.2,0.4,0.6,0.8,1.0},
      ytick={0,0.2,0.4,0.6,0.8,1.0},
      tick align=outside,
      major tick length=2.5pt,
      axis line style={line width=0.5pt},
    },
    robotcycle/.style={
      cycle list={
        {green!55!black,solid},
        {violet,solid},
        {orange!85!black,solid},
        {blue,solid}
      }
    },
    thinlines/.style={
      every axis plot/.append style={line width=0.75pt}
    }
  }

  \begin{groupplot}[
    group style={
        group size=2 by 2, 
        horizontal sep=0.35cm, 
        vertical sep=0.9cm
    },
    width=0.38\textwidth, 
    height=0.38\textwidth,
    baseaxis, 
    robotcycle, 
    thinlines,
  ]

  \nextgroupplot[
    ylabel={Precision},
    xticklabels={},
    title={\textbf{Kittredge Loop Multi-Robot}},
    title style={yshift=-5pt},
    legend to name=PRlegend,
    legend columns=4
  ]
    \foreach \r in {robot1,robot2,robot3,robot4}{
        \addplot+[line width=0.9pt] table[x=recall, y=\r, col sep=comma]{figures/data/PR_KL_multi_robot.csv};
    }
    \addlegendentry{\textit{robot1}};
    \addlegendentry{\textit{robot2}};
    \addlegendentry{\textit{robot3}};
    \addlegendentry{\textit{robot4}};

  \nextgroupplot[
    ylabel={}, 
    xticklabels={}, 
    yticklabels={},
    title={\textbf{Kittredge Loop Multi-Session}},
    title style={yshift=-5pt},
  ]
    \foreach \r in {robot1,robot2,robot3,robot4}{
        \addplot+[line width=0.9pt] table[x=recall, y=\r, col sep=comma]{figures/data/PR_KL_multi_session.csv};
    }

  \nextgroupplot[
    xlabel={Recall},
    ylabel={Precision},
    title={\textbf{Main Campus Multi-Robot}},
    title style={yshift=-5pt},
  ]
    \foreach \r in {robot1,robot2,robot3,robot4}{
        \addplot+[line width=0.9pt] table[x=recall, y=\r, col sep=comma]{figures/data/PR_MC_multi_robot.csv};
    }
    
  \nextgroupplot[
    xlabel={Recall},
    ylabel={},
    yticklabels={},
    title={\textbf{Main Campus Multi-Session}},
    title style={yshift=-5pt},
  ]
    \foreach \r in {robot1,robot2,robot3,robot4}{
        \addplot+[line width=0.9pt] table[x=recall, y=\r, col sep=comma]{figures/data/PR_MC_multi_session.csv};
    }

  \end{groupplot}

  \node[anchor=center, overlay] at
    ($(group c1r2.north)!0.5!(group c2r2.south)+(-8em,-11.3em)$)
    {\pgfplotslegendfromname{PRlegend}};
\end{tikzpicture}
}
  \caption{Precision-Recall curves for the ScanContext descriptor for LiDAR-based place recognition for both environments in our dataset using a multi-session test and a multi-robot test.}
  \label{fig:LPR_precision_recall}
\end{figure}

\subsection{LiDAR-Based C-SLAM}
DiSCo-SLAM exchanges compact ScanContext descriptors and performs a two-stage global–local pose-graph optimization for distributed multi-robot SLAM. The reference implementation supports up to three robots, so we extended it to handle all four robots available in CU-Multi. We tested DiSCo-SLAM four times in the Main Campus environment and five times in the Kittredge Loop environment. Of these trials, one run on Main Campus and two runs on Kittredge Loop diverged due to erroneous loop closures. To ensure that inter-robot loop closures did not occur immediately, playback was offset by 200 seconds from the initial start time. We report the averages for Absolute Trajectory Error (ATE) and Relative Pose Error (RPE) on all runs that remained stable and did not fail due to incorrect loop closure detections (Table~\ref{tab:discoslam_results}). Figure~\ref{fig:discoslam_full} illustrates an example of progressive inter-robot alignment produced by DiSCo-SLAM across both environments.

\begin{table}[ht]
    \caption{DiSCo-SLAM results on the CU-Multi Dataset}
    \label{tab:discoslam_results}
    \centering
    \begin{tabular}{llrrrr} 
        \toprule
        \textbf{Environment} & & \textbf{robot1} & \textbf{robot2} & \textbf{robot3} & \textbf{robot4} \\ [0.5ex]
        \midrule 
        Main Campus & \emph{ATE (m)} & 5.58 & 8.52 & 14.30 & 22.85\\
        & \emph{RPE (m)} & 0.11 & 0.11 & 0.49 & 0.31\\
        \midrule 
        Kittredge Loop & \emph{ATE (m)} & 7.68 & 5.01 & 14.42 & 18.85\\
        & \emph{RPE (m)} & 0.08 & 0.16 & 0.35 & 0.45\\
        \bottomrule
    \end{tabular}
    \\[0.5ex]
    \footnotesize{}
\end{table}

\subsection{Results}
From our experiments, we found that on high-overlap pairs (e.g., \textit{robot1} and \textit{robot2}), DiSCo-SLAM rapidly establishes inter-robot loop closures leading to better overall ATE and ScanContext yields stronger precision-recall. As overlap decreases (\textit{robot1} and \textit{robot4}), PR degrades and time-to-first closure increases, quantifying the practical effect of viewpoint diversity and partial overlap that CU-Multi was designed to expose. These results, together with our globally aligned ground truth and semantics, showcase CU-Multi as the right substrate for realistic evaluation without resorting to single-trajectory splitting.

\section{Conclusion and Future Work}
We introduced CU-Multi, a multi-robot dataset that addresses persistent challenges in evaluating collaborative perception and SLAM. By combining diverse trajectories with controlled overlap, synchronized multi-modal sensing, semantic LiDAR annotations, and refined ground-truth odometry across two large-scale outdoor environments, CU-Multi enables realistic and reproducible experimentation beyond single-robot trajectory partitioning. Baseline evaluations confirm that varied overlap levels expose meaningful performance differences in both place recognition and C-SLAM. The public release of the dataset, support code, and documentation is intended to accelerate progress in collaborative autonomy. In future work, we plan to extend CU-Multi to include heterogeneous platforms and indoor environments.


\bibliographystyle{IEEEtran}
\bibliography{references}

\end{document}